\documentclass[letterpaper, 10 pt, journal, twoside]{ieeetran}

\usepackage{graphics} 
\usepackage{amsmath} 
\usepackage{amssymb}  
\usepackage{booktabs}

\usepackage{multirow}
\usepackage{tabularx}
\usepackage{graphicx}
\usepackage{adjustbox}
\usepackage{subcaption}
\usepackage{xcolor}
\usepackage[font=footnotesize]{caption}
\allowdisplaybreaks

\usepackage{pifont}
\newcommand{\cmark}{\textcolor{green}{\ding{51}}} 
\newcommand{\xmark}{\textcolor{red}{\ding{55}}}   

\usepackage{booktabs}  
\newcommand{\stdgray}[1]{\textcolor{gray}{\scriptsize(±#1)}}
\newcommand{\meanstd}[2]{#1\;\stdgray{#2}}

\newif\ifrevmarks
\revmarksfalse
\ifrevmarks
  \usepackage[authormarkup=none]{changes}
  \newcommand{\shmath}[1]{\textcolor{blue}{#1}}
\else
  \usepackage[final]{changes}
  \newcommand{\shmath}[1]{#1}
\fi
\setdeletedmarkup{}
\definechangesauthor[name={Morgan}, color=red]{MB}
\definechangesauthor[name={Sehoon}, color=blue]{SH}

\title{\LARGE \bf
AdaptManip: Learning Adaptive Whole-Body Object Lifting and
Delivery with Online Recurrent State Estimation
}

\author{Morgan Byrd$^{1*}$, Donghoon Baek$^{1}$, Kartik Garg$^{1}$, Hyunyoung Jung$^{1}$,

Daesol Cho$^{1}$, Maks Sorokin$^{1}$, Robert Wright$^{2}$, and Sehoon Ha$^{1}$
\thanks{$^{1}$Georgia Institute of Technology, Atlanta, GA, 30308, USA}%
\thanks{$^{2}$Georgia Tech Research Institute, Atlanta, GA, 30308, USA}%
\thanks{*Correspondence to abyrd45@gatech.edu}
}

\begin{document}

\maketitle

\begin{abstract}
This paper presents Adaptive Whole-body Loco-Manipulation, AdaptManip, a fully autonomous framework for humanoid robots to perform integrated navigation, object lifting, and delivery. Unlike prior imitation learning-based approaches that rely on human demonstrations and are often brittle to disturbances, AdaptManip aims to train a robust loco-manipulation policy via reinforcement learning without human demonstrations or teleoperation data. 
The proposed framework consists of three coupled components: (1) a recurrent object state estimator that tracks the manipulated object in real time under limited field-of-view and occlusions; (2) a whole-body base policy for robust locomotion with residual manipulation control for stable object lifting and delivery; and (3) a LiDAR-based robot global position estimator that provides drift-robust localization. All components are trained in simulation using reinforcement learning and deployed on real hardware in a zero-shot manner.  
Experimental results show that AdaptManip significantly outperforms baseline methods, including imitation learning-based approaches, in adaptability and overall success rate, while \replaced[id=SH]{the learned estimator keeps tracking the object when visual observations are intermittent}{accurate object state estimation improves manipulation performance even under occlusion}. We further demonstrate fully autonomous real-world navigation, object lifting, and delivery on a humanoid robot.

\end{abstract}


\section{INTRODUCTION}
\label{S:1}









Humanoid robots are a promising platform for human-centric environments due to their ability to execute human-like whole-body capabilities \cite{bostondynamics_atlas, tesla_optimus_bi}. 
However, achieving reliable whole-body loco-manipulation on humanoids remains fundamentally challenging. 
Such tasks require the robot to simultaneously coordinate high-dimensional whole-body dynamics, maintain balance under changing contact conditions, and regulate complex multi-contact interactions with external objects. While recent work \cite{yin2025visualmimic,zhao2025resmimic,yang2025omniretarget,chen2025gmt} has enabled dynamic whole-body skills such as jumping and parkour, autonomous contact-rich humanoid whole-body loco manipulation remains largely unsolved.

To address these challenges, recent advances in imitation learning—leveraging motion capture demonstrations, teleoperation, and large-scale human video datasets—have emerged as a powerful paradigm for learning whole-body humanoid control, which offers an alternative to traditional model-based approaches \cite{chen2025gmt, ze2025twist, he2024omnih2o, he2026ultra}. While such methods have shown strong performance, they rely heavily on motion capture systems and curated demonstrations that provide privileged global robot and object state information \cite{weng2025hdmi,yin2025visualmimic, zhao2025resmimic, yang2025omniretarget}. Moreover, recovery and adaptability under natural failure cases, such as object slippage or drops, remain underexplored because these methods tend to follow the fixed reference motion for the given time window.

\begin{figure}[!t]
    \centering
    \includegraphics[width=0.85\columnwidth]{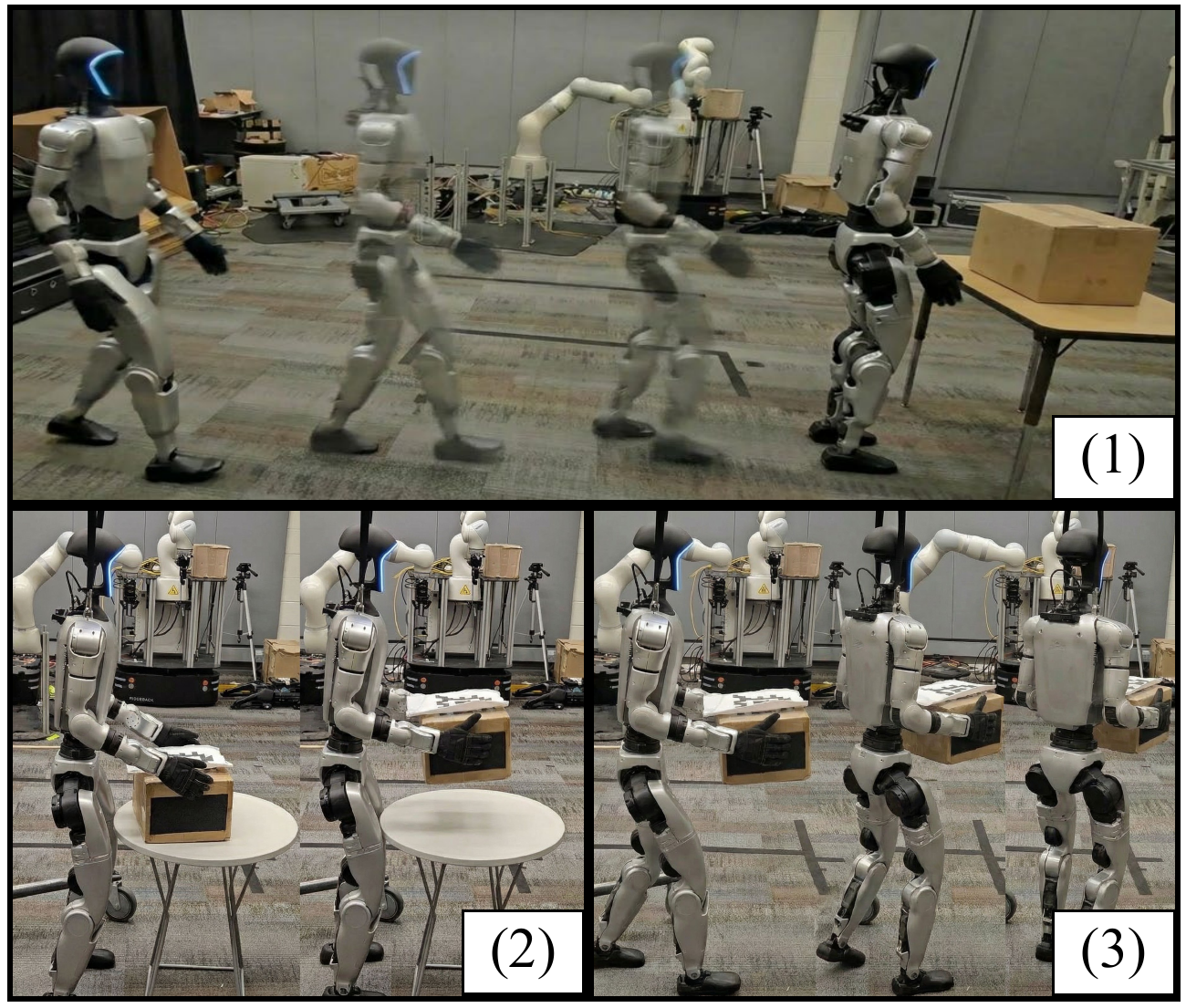}
    \caption{\textbf{Fully autonomous humanoid loco-manipulation using online recurrent state estimation.} (1) navigating toward the object, (2) lifting the object through coordinated whole-body motion, and (3) delivering the object to the target location. Our method relies solely on onboard sensing and does not require teleoperation data or an external mocap system.}
    \label{fig1}
\vspace{-1.5em}
\end{figure}

Our key insight for robust and adaptive loco-manipulation is to \textit{jointly learn online object state estimation and control by training a recurrent state estimator alongside reinforcement learning that encounters and adapts to failure cases.} 
This formulation tightly couples perception and control, and allows the humanoid to continuously reason about object pose and contact evolution without relying on privileged global information or pre-recorded demonstrations. 
By integrating object state estimation into training, the robot naturally learns to adjust its whole-body coordination when interactions deviate from nominal execution, such as during object slippage or partial loss of contact. 
Inspired by human behavior, our design leverages complementary sensing modalities—vision and proprioception—to enable robust state inference even when visual observations are partial or occluded.

Motivated by this insight, we propose AdaptManip, a learning-based framework for whole-body humanoid loco-manipulation that jointly integrates online object state estimation and control to enable fully autonomous navigation, lifting, and delivery using only onboard sensing. By coupling multi-modal object state estimation from vision and proprioception with LiDAR-based robot pose odometry, AdaptManip achieves robust and recovery-capable loco-manipulation—including regrasping after object drops—without relying on motion capture systems or teleoperation. See Table~\ref{table:comparison} for a qualitative comparison to existing methods. To realize these capabilities, our framework integrates three key components into a unified system: (1) a reinforcement learning-based locomotion policy for stable bipedal mobility, (2) a residual upper-body manipulation policy for contact-rich object interaction, and 
(3) a fully onboard object state estimator that provides real-time perception for control. Concretely, the robot operates through three coordinated stages: navigation, lifting, and delivery. 
During the navigation stage, the humanoid approaches the target object using LiDAR-based robot pose odometry and proprioceptive feedback, enabling fully onboard localization. In the lifting stage, an online object state estimator guides grasping and coordinated whole-body manipulation. In the delivery stage, the robot transports the object to the target location while maintaining balance under changing contact conditions. Overall, AdaptManip achieves higher task success rates than prior baselines by learning robust and adaptive behaviors through reinforcement learning, including recovery actions under failure conditions.
This is realized in a fully autonomous system through the integration of onboard sensing and recurrent state estimation.

We summarize our key contributions as follows. 
First, we introduce AdaptManip, a learning-based framework for whole-body humanoid loco-manipulation that autonomously accomplishes navigation, lifting, and delivery through a structured three-stage strategy. 
Second, we develop an online, recurrent object state estimation module that fuses vision and proprioceptive sensing, enabling robust and recovery-capable loco-manipulation using only onboard sensors, without teleoperation data or external motion capture systems. 
Finally, we validate the effectiveness of AdaptManip through extensive simulation studies and real-world experiments on physical humanoid hardware.

\section{RELATED WORK}
\label{Relatedwork}

\begin{table}[t] 
\centering
\caption{Comparison of representative methods across key aspects:
onboard-only sensing (Onbd), absence of human demonstrations (NoHumRef),
locomotion–manipulation capability (LocoMan),
absence of future references (NoFutRef),
and no teleoperation (NoTeleOp).}
\resizebox{\columnwidth}{!}{
    \begin{tabular}{lccccc}
    \toprule
    \textbf{Method} &
    \textbf{Onbd} &
    \textbf{NoHumRef} &
    \textbf{LocoMan} &
    \textbf{NoFutRef} &
    \textbf{NoTeleOp} \\
    \midrule
    TWIST \cite{ze2025twist}        & \xmark & \xmark & \cmark & \cmark & \cmark \\
    ResMimic \cite{zhao2025resmimic}     & \xmark & \xmark & \cmark & \cmark & \cmark \\
    VisualMimic \cite{yin2025visualmimic}  & \cmark & \xmark & \cmark & \cmark & \xmark \\
    HDMI \cite{weng2025hdmi}         & \xmark & \xmark & \cmark & \cmark & \cmark \\
    PhysHSI \cite{wang2025physhsi}     & \xmark & \xmark & \cmark & \cmark & \cmark \\
    OmniRetarget \cite{yang2025omniretarget} & \xmark & \xmark & \cmark & \cmark & \cmark \\
    GMT \cite{chen2025gmt}         & \cmark & \xmark & \cmark & \xmark & \cmark \\
    BoxLocoManip \cite{dao2024sim}  & \cmark & \cmark & \cmark & \cmark & \xmark \\
    \midrule
    \textbf{Ours} & \cmark & \cmark & \cmark & \cmark & \cmark \\
    \bottomrule
    \end{tabular}%
}
\label{table:comparison}
\vspace{-1em}
\end{table}

\subsection{Humanoid Whole-Body Control} 

Over the past decades, model-based control, such as Model Predictive Control, has advanced significantly across a wide range of humanoid platforms, including quadrupedal, bipedal, and wheeled robots \cite{Kuindersma2016, donghyun_tro, westervelt2003hybrid,baek2025whole}. These approaches are typically developed through hierarchical and decomposed control architectures that separate trajectory planning and whole-body control, often integrating model-based predictive optimization at different levels and time scales \cite{li2024cafe, khazoom2024tailoring, kim1909highly}. A key advantage of this paradigm is its reliance on physics-based models, which provide strong interpretability, stability guarantees, and smooth, dynamics-aware continuous control actions. However, the performance is inherently sensitive to assumptions and often requires manual human effort to model accurate loco-manipulation behaviors.

In light of these challenges, learning-based methods
—particularly reinforcement learning (RL)—
have become increasingly influential in humanoid robotics, supporting the synthesis of complex whole-body behaviors~ \cite{kormushev2013reinforcement}. End-to-end RL approaches, often augmented with domain randomization, have shown encouraging sim-to-real transfer and demonstrated successful deployment on hardware platforms \cite{tan2018sim, rho2025unsupervised, lee2020learning, kim2025understanding, kumar2021rma, jung2025ppf}. However, reliable and fully autonomous object-centric whole-body manipulation remains challenging, as physical interaction with diverse objects entails complex, task-dependent dynamics that are difficult to model and generalize.

\vspace{-0.65em}
\subsection{Learning Humanoid Loco-Manipulation} 
Learning-based methods have recently advanced from isolated locomotion and manipulation to integrated whole-body humanoid loco-manipulation~\cite{he2026ultra, chen2025gmt, zhang2025any2track, xie2025kungfubot}. This progress has been accelerated by the availability of large-scale open-source motion datasets (e.g., AMASS \cite{AMASS} and LAFAN1 \cite{harvey2020robust}), together with advances in imitation learning, which enable humanoid robots to reproduce natural human motions such as jumping, running, dancing, and kicking \cite{chen2025gmt,zhang2025any2track,kim2025switch,xie2025kungfubot,li2025clone}. Despite these successes, existing imitation-based humanoid controllers largely emphasize kinematic motion reproduction, with limited ability to handle object contacts and interaction dynamics.

Recent works have begun to address object-centric whole-body humanoid loco-manipulation, where robots must jointly reason about physical interaction with objects. Existing approaches explore this problem using motion imitation with external tracking \cite{weng2025hdmi,zhao2025resmimic}, vision-based reinforcement learning \cite{yin2025visualmimic, he2026ultra}, or hybrid perception pipelines combining long-range sensing and close-range visual feedback \cite{wang2025physhsi}. While these methods demonstrate promising results on tasks such as box lifting, many rely on strong assumptions, including motion capture supervision during inference, human-in-the-loop navigation or skill transitions \cite{dao2024sim}, and continuous visual access to the target object. Unlike existing works, our approach relies solely on fully onboard sensing and enables fully autonomous, robust, and adaptive whole-body behaviors to accomplish contact-rich manipulation tasks.

\vspace{-0.65em}
\subsection{Object State Estimation}
Accurate perception and state estimation are essential and nontrivial for humanoid loco-manipulation tasks.
Motion capture systems are commonly used to obtain accurate global robot and object poses in laboratory environments \cite{weng2025hdmi, zhao2025resmimic}. However, such systems are inherently restricted to controlled environments and are impractical for deployment beyond the lab. Visual pose estimation offers a more flexible alternative, either through fiducial markers \cite{olson2011tags} or direct pixel-based methods \cite{foundationposewen2024}, enabling operation in less structured settings. Despite their effectiveness, vision-based approaches typically require continuous and reliable visual observations, making them vulnerable to failures caused by occlusions, limited view frustums, or dynamic viewpoints during whole-body motion. Other egocentric works like \cite{seo2023trill, fu2024humanplus} rely on teleoperation and modified hardware for improved field of view and reliability. We rely exclusively on fully onboard sensing with nominal hardware—camera, LiDAR, and robot proprioception—to, similar to \cite{byrd-2025}, recurrently estimate object state, robot–object contact forces, and the robot’s global pose, and deploy the resulting system directly on hardware for fully autonomous loco-manipulation.

\vspace{-0.65em}
\section{Task Description}
\label{task_depscription}
\begin{figure}[t]
    \centering
    \includegraphics[width=\columnwidth]{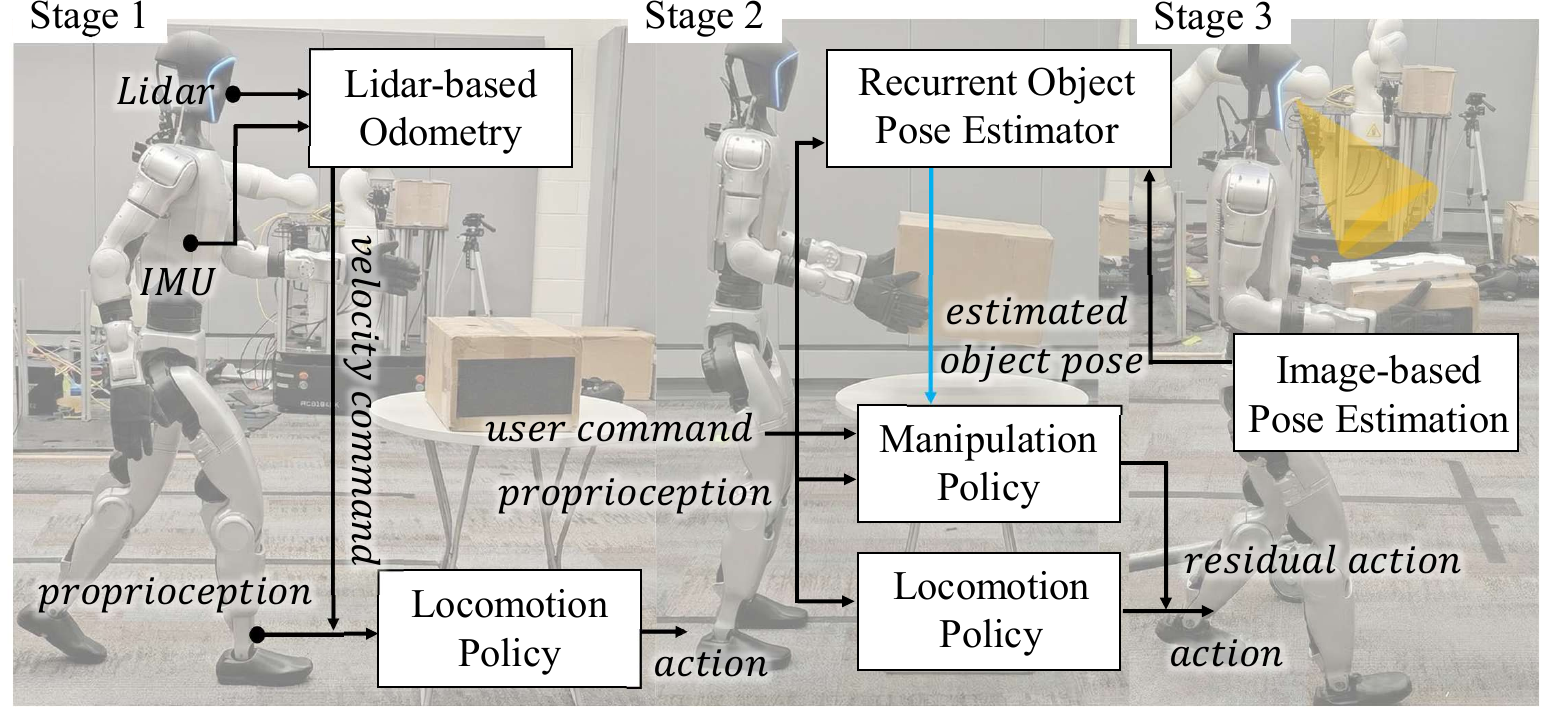}
    \caption{\textbf{Three-stage AdaptManip experiment plan and deployment.} 
    Stage~1: LiDAR odometry and proprioception enable autonomous navigation.
    Stage~2: Recurrent multimodal object-pose estimation supports coordinated lifting.
    Stage~3: Image-based refinement and residual policies ensure stable delivery.
    All stages operate using only onboard sensing.}
    \label{fig:overview_exp}
\vspace{-1em}
\end{figure}

To robustly complete our whole-body loco-manipulation tasks, we decompose it into three stages: (1) moving the robot from an initial pose to the box, (2) grasping and lifting the box, and (3) transporting it to the target location. Each stage is designed to address the distinct requirements (Fig.~\ref{fig:overview_exp}).

\noindent\textbf{Navigation.} 
In the first stage, the humanoid navigates from its initial location to a goal location in front of the target object, which is given as input. Therefore, this stage requires locomotion rather than object manipulation skills. Because the object is initially outside the camera’s field of view, the relative object location is estimated using LiDAR-inertial odometry based on the onboard LiDAR sensor, implemented with FAST-LIO \cite{fastlio}. While we use LiDAR for localization due to the availability on our hardware, any other equivalently performing localization method would work.

\noindent\textbf{Grasping and Lifting.}
When the robot approaches sufficiently close to the object ($\leq 0.5$m), it transitions to the second stage to grasp and lift the target object. The robot employs unarticulated rubber hands; therefore, the object is supported purely through frictional contact rather than form closure. In this phase, vision-based sensing is activated to refine the object pose for accurate grasping and lifting. We implement the vision-based sensing using an AprilTag~\cite{olson2011tags}, although it can be replaced with any vision-based 6D object pose estimator.
An effective policy for this stage must be robust to inaccurate or unavailable 6D object pose information.



\noindent\textbf{Carrying to Destination.}
Once the object is securely lifted, the system transitions to the third stage, which involves transporting the object from its initial location to the target location. During this phase, the object is frequently occluded by the robot’s body and hands and visual information becomes unreliable. 

\vspace{-0.65em}
\section{Learning Loco-manipulation \\ with Recurrent State Estimation}
\label{M:1}

\begin{figure*}[!t]
\centering
\includegraphics[width=16cm]{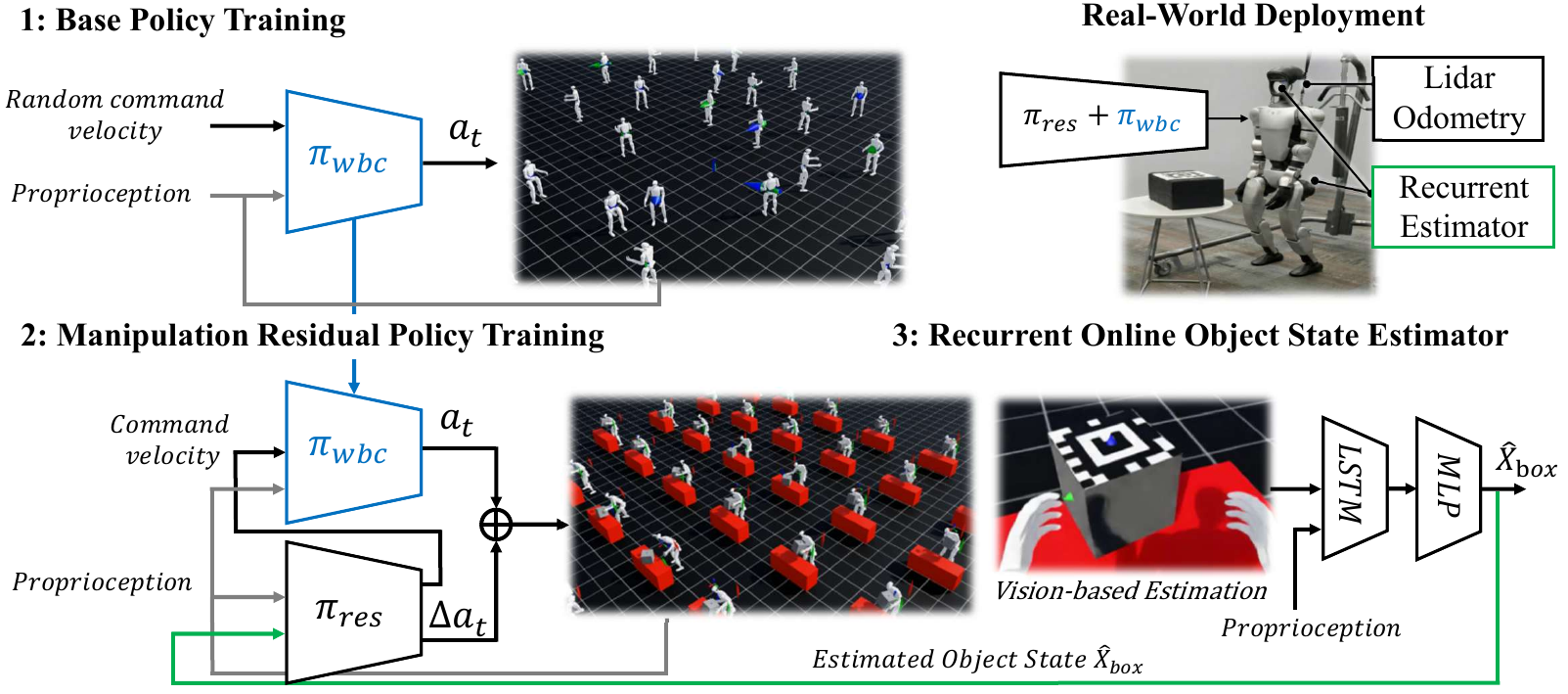}
\caption{\textbf{Overview of the training and deployment pipeline.} 
(1) A base whole-body control policy $\pi_{\mathrm{wbc}}$ is trained in IsaacLab to generate base whole-body behavior such as walking. 
(2) A manipulation residual policy $\pi_{\mathrm{res}}$ is trained on top of the base policy, taking proprioception and the estimated object state $\hat{X}_{\mathrm{box}}$ to produce residual actions $\Delta a_t$. The residual action aims to adaptively lift a 3D object. 
(3) A recurrent online object state estimator fuses vision and proprioceptive cues using a V-LSTM and MLP to infer $\hat{X}_{\mathrm{box}}$, and is trained jointly with the residual manipulation policy.
During real-world deployment, the robot uses onboard estimators and LiDAR odometry and executes the combined policies $\pi_{\mathrm{wbc}}$ and $\pi_{\mathrm{res}}$ to complete the whole-body loco-manipulation task. 
\vspace{-1em}
}


\label{fig:overview_of_adaptmanip}
\end{figure*}

We present a learning-based humanoid loco-manipulation policy with a recurrent object state estimator to enable robust object grasping, lifting, and delivery under unreliable visual observations. The proposed policy allows a humanoid robot to adaptively manipulate an object by integrating onboard visual and proprioceptive information. However, learning such a policy is inherently difficult, as humanoid loco-manipulation requires simultaneously maintaining stable bipedal locomotion, interpreting noisy and intermittent visual inputs, and executing adaptive whole-body manipulation under contact-rich interactions. This challenge substantially increases the complexity of policy learning and reduces the effectiveness of a naive end-to-end approach.

To address these challenges, we structure the learning process into three components. First, we train a base whole-body locomotion policy that provides stable and periodic bipedal walking. Then, we learn a whole-body residual manipulation policy on top of the base policy, which enables adaptive object grasping and lifting while preserving locomotion stability. Concurrently, we train a recurrent object state estimator that fuses visual observations and proprioceptive signals to infer the object pose online, which aims to achieve robust manipulation under partial or missing visual inputs, similar to \cite{ji2022concurrent}. 
The details of each stage are described in the following subsections.


\vspace{-0.65em}
\subsection{Base Whole-Body Locomotion Policy}
For robust whole-body walking, we train a base locomotion policy using RL to generate stable bipedal locomotion across a range of commanded velocities. This policy serves as a fixed foundation for subsequent whole-body manipulation learning.

\noindent\textbf{Observation and Action Space.}
We employ an asymmetric actor--critic architecture during training. The actor observation at time $t$ is defined as
\begin{equation*}
    \mathbf{o}_t^{\text{actor}} =
\left[
\; \boldsymbol{\theta}_j,\,
\dot{\boldsymbol{\theta}}_j,\,
\boldsymbol{\omega},\,
\mathbf{g}_{\mathrm{proj}},\,
\mathbf{\bar{v}},\,
\bar{\omega}_{z},\,
\bar{z},\,
\mathbf{a}_{t-1}
\;\right],
\label{base_policy_obs}
\end{equation*}
where $\boldsymbol{\theta}_j \in \mathbb{R}^{29}$ and $\dot{\boldsymbol{\theta}}_j \in \mathbb{R}^{29}$ denote the joint positions and velocities, $\boldsymbol{\omega} \in \mathbb{R}^{3}$ is the base angular velocity, $\mathbf{g}_{\mathrm{proj}} \in \mathbb{R}^{3}$ is the gravity vector projected into the torso frame, $\mathbf{\bar{v}} = [\bar{v}_x, \bar{v}_y]^\top \in \mathbb{R}^{2}$ denotes the commanded planar velocity of the base, $\bar{\omega} \in \mathbb{R}$ is the commanded base yaw rate, $\bar{z} \in \mathbb{R}$ is the commanded base height, and $\mathbf{a}_{t-1} \in \mathbb{R}^{29}$ denotes the action applied at the previous timestep. The critic receives the same observations as the actor, augmented with the base linear velocity $\mathbf{v}_t \in \mathbb{R}^3$, and is provided with a three-step temporal history of all inputs, i.e., $\mathbf{o}_t^{\text{critic}} = [\,\mathbf{o}_{t-2:t}^{\text{actor}},\; \mathbf{v}_{t-2:t}\,]$.


The actions for the policy, $\mathbf{a}_t \in \mathbb{R}^{29}$, are joint-space position targets relative to a default configuration which are mapped to joint torques via a PD controller, $\boldsymbol{\tau}_t = \mathbf{k}_p \bigl((\mathbf{a}_t + \mathbf{q}_{\mathrm{def}}) - \mathbf{q}_t\bigr) - \mathbf{k}_d \dot{\mathbf{q}}_t$.

\noindent\textbf{Reward Design.}
\label{sec:reward_locomotion}
The locomotion reward follows a weighted structure consisting of
command tracking, gait shaping, motion regularization,
and constraint violation penalties, as summarized in Eq. (\ref{base_policy_reward_eqn}),
following prior work~\cite{pmlr-v164-rudin22a}.

{\small
\begin{align}
r_{\text{loco}}
=&\;
\underbrace{
\shmath{\boldsymbol{\lambda}_{\text{tr}}\cdot\!\big[
e^{-\|v-\bar{v}\|},\,
e^{-\|\omega-\bar{\omega}\|},\,
e^{-|z-\bar{z}|}
\big]}
}_{\text{Command tracking}}
\nonumber\\
+
&\underbrace{
\shmath{\boldsymbol{\lambda}_{\text{gait}}\cdot\!\big[
\sum_f \mathbb{I}_{v>0.1}(t_f-0.4),\,
e^{-0.5\sum_f|z_f-0.05|}
\big]}
}_{\text{Gait shaping}}
\nonumber\\
\displaybreak[3]
-
&\underbrace{
\shmath{\boldsymbol{\lambda}_{\text{reg}}\cdot
\sum_j \big[
\tau_j^2,\,
\omega_j^2,\,
\dot{\omega}_j^2,\,
a_j^2,\,
\dot{a}_j^2,\,
\ddot{a}_j^2
\big]}
}_{\text{Motion regularization}}
\nonumber\\
-
&\underbrace{
\shmath{\boldsymbol{\lambda}_{\text{vio}}\cdot\!\big[
\sum_f \mathbb{I}_c \|v_f\|,\,
\sum_f v_r f_c,\,
\|g_\mathrm{proj}\|^2,\,
\omega_o,\,
\tau_o
\big]}
}_{\text{Constraint violation penalties}} .
\label{base_policy_reward_eqn}
\end{align}
}

\added[id=SH]{Each $\boldsymbol{\lambda}$ collects the per-term weights of its group.} Here, $v$, $\omega$, and $z$ denote the base linear velocity, angular velocity, and height, with corresponding commands $\bar{v}$, $\bar{\omega}$, and $\bar{z}$. The index $j$ denotes joints and $f$ denotes feet. Terms $\tau_j$, $\omega_j$, and $\dot{\omega}_j$ are joint torque, velocity, and acceleration, while $a_j$ is the policy action. $\dot{a}_j$ and $\ddot{a}_j$ denote the first- and second-order finite differences of actions.
$v_f$ and $f_c$ are the foot velocity and contact force, $t_f$ is the foot air-time, and $z_f$ is the foot height. $g_\mathrm{proj}$ denotes the gravity vector projected onto the body frame. $\omega_o$ and $\tau_o$ indicate joint velocity and torque limit violations. $\mathbb{I}_c$ is an indicator function for foot contact. 


\subsection{Whole-Body Residual Manipulation Policy}
For whole-body loco-manipulation, we train a \textbf{residual} policy on top of the frozen base policy. This residual policy learns task-specific adaptations for object grasping, lifting, and stabilization.

\noindent\textbf{Observation and Action Space.} The actor receives
\begin{equation*}
\mathbf{o}_t^{\text{actor}} =
\left[
\boldsymbol{\theta}_j,\,
\dot{\boldsymbol{\theta}}_j,\,
\boldsymbol{\omega},\,
\mathbf{g}_{\mathrm{proj}},\,
\tilde{\mathbf{X}}_{\mathrm{box}},\,
\bar{\mathbf{X}}_{\mathrm{box}},\,
\mathbf{a}_{t-1}
\right],
\end{equation*}
which includes the same proprioceptive state as the base locomotion policy (joint positions $\boldsymbol{\theta}_j$, velocities $\dot{\boldsymbol{\theta}}_j$, base angular velocity $\boldsymbol{\omega}$, and projected gravity $\mathbf{g}_{\mathrm{proj}}$), augmented with the estimated $6$D box pose $\tilde{\mathbf{X}}_{\text{box}}$, the commanded box pose $\bar{\mathbf{X}}_{\text{box}}$, and the previous action $\mathbf{a}_{t-1}$. As in the locomotion controller, we employ an asymmetric actor–critic architecture. The critic augments the actor observations with privileged information:
\begin{equation*}
\mathbf{o}_t^{\text{critic}} =
\left[
\mathbf{o}_{t-2:t}^{\text{actor}},\,
\mathbf{o}_{t-2:t}^{\text{priv}}
\right],
\end{equation*}
where $\mathbf{o}_t^{\text{priv}}$ denotes privileged information available only during training, including 
the ground-truth 6D pose $\mathbf{X}_\mathrm{box}$, 
the linear velocity $\mathbf{v}_\mathrm{box}$, the angular velocity $\mathbf{\omega}_\mathrm{box}$, as well as hand and box contact forces $\mathbf{f}_{\mathrm{hand}}$ and $\mathbf{f}_{\mathrm{box}}$.


The action space consists of two components.
First, the command vector provided to the frozen low-level locomotion policy is
$\bar{\mathbf{u}}_{\text{loco}} = [\bar{v}_{x}, \bar{v}_{y}, \bar{\omega}, \bar{z}] \in \mathbb{R}^{4}$,
which corresponds to the base motion and height commands defined in (\ref{base_policy_obs}).
These commands regulate planar locomotion, including forward and lateral motion, turning, and base height control.
Second, the upper-body action $a_{\text{upper}} \in \mathbb{R}^{17}$ corresponds to residual PD joint position targets for the upper body, primarily affecting the waist and arm joints.

\noindent\textbf{Reward Design.}
We use the locomotion reward $r_{\text{loco}}$ defined in
Eq.~(\ref{base_policy_reward_eqn})
and augment it with manipulation-specific objectives for robust bimanual
grasping and stable box transport.
Beyond the base locomotion reward in
Eq.~(\ref{base_policy_reward_eqn}),
the proposed reward introduces additional manipulation-specific objectives
that explicitly account for physical interaction with the object during
bimanual grasping and transport.
In particular, the contact-related terms penalize excessive relative motion
between the robot and the box, encourage symmetric bimanual contact forces,
reward proper hand--box contact orientation, penalize failures to establish
contact, and discourage tangential hand motion indicative of slipping:

{\small
\begin{align}
r
=&\; r_{\text{loco}}
\nonumber\\
+
&\underbrace{
\shmath{\boldsymbol{\lambda}_{\text{kin}}\cdot\!\Big[
e^{-|\psi_{\text{robot}}-\psi_{\text{box}}|},\,
e^{-4\|p_{\text{hand}}^{\text{err}}\|},\,
e^{-1.5\|p_{\text{root}}^{\text{err}}\|}
\Big]}
}_{\text{Kinematic tracking}}
\nonumber\\
+
&\underbrace{
\shmath{\boldsymbol{\lambda}_{\text{box}}\cdot\!\Big[
e^{-2\|p_{\text{box}}-p_{\text{des}}\|_1-\|q_{\text{box}}-q_{\text{des}}\|_1},\,
e^{-\|v_{\text{root}}-v_{\text{box}}\|_2}
\Big]}
}_{\text{Box stabilization}}
\nonumber\\
\displaybreak[3]
+
&\underbrace{
\shmath{\lambda_{\text{con}}}\,
\text{clamp}\!\Big(
\sum_h \|f_{\text{con},h}\|\,\mathbb{I}_{\text{box}},\,0,\,1
\Big)
}_{\text{Contact force quality}}
\nonumber\\
-
&\underbrace{
\shmath{\lambda_{\text{slip}}}
\sum_h \min\!\big(0,\, v_{\text{hand},z}-v_{\text{box},z}\big)
}_{\text{Slip avoidance}} .
\label{eq:residual_reward}
\end{align}
}

\noindent\textbf{Terminal Condition.}
Episodes terminate after 20 seconds, if the robot tilts more than $60^\circ$, if the root height drops below $0.15\mathrm{m}$, or if the box falls below $0.25\mathrm{m}$.


\subsection{Recurrent Object State Estimation}
Accurate and robust object state estimation is essential for whole-body loco-manipulation, as manipulation performance directly depends on reliable object pose information. Although motion capture systems provide accurate measurements, they are restricted to laboratory settings, making vision-based perception essential for real-world operation. In this work, we employ a robust tag-based visual pose estimator~\cite{olson2011tags}; however, visual observations are often incomplete or intermittent due to occlusions, limited camera field of view, and object motion during manipulation. Enforcing constant object visibility to mitigate these issues can induce unnatural behaviors. 

We propose an online object state estimation approach that fuses visual observations with proprioceptive measurements and executed actions. The estimator follows a human-inspired strategy, using vision primarily for grasp initiation and maintaining the object state through proprioception during manipulation. By leveraging proprioceptive and action histories, the estimator remains reliable under partial or intermittent visual feedback.

We employ a recurrent object state estimator to infer the relative object pose $\tilde{\mathbf{X}}_{\mathrm{box}}$ online during manipulation. The estimator takes as input the current vision-based pose measurement (set to zero when unavailable), robot proprioceptive observations, and the executed action, and outputs the estimated $6$D pose in the robot frame. We implement the estimator using an LSTM with an MLP output head, as its internal memory enables robust state propagation under missing or unreliable visual observations.

The estimator is trained concurrently with the manipulation policy using supervised learning with the ground-truth object poses available in simulation. To stabilize early training, we apply a curriculum that gradually replaces the ground-truth pose with the estimated pose in the policy input. Specifically, the pose provided to the policy is defined as
\begin{equation}
\mathbf{X}_{\text{in}} = w\,\tilde{\mathbf{X}}_{\mathrm{box}} + (1 - w)\,\mathbf{X}_{\mathrm{box}},
\end{equation}
where the weighting factor is given by $w = \min(t / T,\, 1)$ with $t$ denoting the current training iteration and $T$ the maximum iteration.

To improve training efficiency, we do not directly incorporate visual inputs during RL training. Instead, we provide the policy with a noisy and randomly masked ground-truth object pose to model visual estimation errors and occlusions, which transfers robustly to real hardware.

\subsection{Domain Randomization}
We employ domain randomization~\cite{Sim2Real2018} during training to improve robustness and sim-to-real transfer by randomizing physical and control parameters, including base mass, ground friction, PD gains, and external disturbances (see Table~\ref{table:manip-randomization}). For whole-body manipulation, we further introduce grasp-specific randomizations, such as box--table friction and restitution, box mass, scale, and center-of-mass location. Observation noise is injected for both the locomotion and manipulation policies to enhance robustness.
\begin{table}[t]
\centering
\begin{tabular}{lll}
\textbf{Parameter} & \textbf{Range} & \textbf{Operation} \\ \hline
Base Mass [kg] & {[}-2.5, 2.5{]} & Add \\
$k_p$ & {[}\ 0.8, 1.2{]} & Scale \\
$k_d$ & {[}\ 0.8, 1.2{]} & Scale \\
Ground Static Friction & {[}\ 0.3, 1.5{]} & Absolute \\
Ground Dynamic Friction & {[}\ 0.3, 0.9{]} & Absolute \\
Base Force Disturbance [N] & {[}-4.0, 4.0{]} & Absolute \\
Base Torque Disturbance [Nm] & {[}-2.0, 2.0{]} & Absolute \\
Table Static Friction & {[}\ 0.3, 1.3{]} & Absolute \\
Table Dynamic Friction & {[}\ 0.3, 1.5{]} & Absolute \\
Table Restitution & {[}\ 0.0, 0.5{]} & Absolute \\
Box Static Friction & {[}\ 0.3, 1.3{]} & Absolute \\
Box Dynamic Friction & {[}\ 0.3, 1.5{]} & Absolute \\
Box Restitution & {[}\ 0.0, 0.5{]} & Absolute \\
Box Mass [kg] & {[}-0.88, 1.5{]} & Add \\
Box Scale $x$ & {[}\ 0.75, 1.25{]} & Scale \\
Box Scale $y$ & {[}\ 0.75, 1.25{]} & Scale \\
Box Center of Mass $x$ & {[}\ 0.75, 1.25{]} & Add \\ 
Box Center of Mass $y$ & {[}\ 0.75, 1.25{]} & Add \\ 
Box Center of Mass $z$ & {[}\ 0.75, 1.25{]} & Add \\ \hline
\end{tabular}
\caption{Domain randomization parameters for policy training.}
\label{table:manip-randomization}
\vspace{-1.85em}
\end{table}

\section{Experimental Results}

In this section, we design simulation and hardware experiments to address the following research questions: (1) Can the proposed method manipulate objects more robustly compared to the baselines? (2) Can the learned object state estimator provide accurate pose estimates? (3) Can our policy be effectively transferred to a real humanoid robot?

\vspace{-0.65em}
\subsection{Implementation Details and Experimental Setup}

\subsubsection{Training Details}
All policies are trained using PPO~\cite{PPO}, augmented with a bilateral symmetry loss~\cite{yu2018learning}. The base locomotion and whole-body manipulation policies are trained separately, each for approximately one day on a single NVIDIA RTX 4090 GPU. The actor and critic are three-layer MLPs with hidden dimensions $[512,\,256,\,128]$ and ELU activations. A recurrent object state estimator, implemented as an LSTM with a hidden dimension of 128, is trained jointly with the policy. 

\subsubsection{Simulation Setup}
We conduct simulation experiments using two physics engines, IsaacLab~\cite{IsaacLab} and MuJoCo~\cite{todorov2012mujoco}, to enable cross-simulator validation. All experiments use a control frequency of 50\,Hz, with the physics simulation running at 200\,Hz. Policies are trained in IsaacLab and evaluated in MuJoCo without additional fine-tuning.

\subsubsection{Hardware Setup}
For real-world evaluation, we deploy on a Unitree G1 humanoid robot~\cite{g1}. The robot is equipped with an Intel RealSense D435i RGB-D camera~\cite{realsense} and a Livox Mid-360 LiDAR~\cite{livox}. All sensing and control are performed onboard, without reliance on external motion capture systems. Using a single AprilTag~\cite{olson2011tags} placed at the center of the top surface of the box, we perform visual object pose estimation using the official Python bindings~\cite{apriltagCode}. The policy is deployed on hardware without additional tuning following sim-to-sim validation, demonstrating effective sim-to-real transfer.

\vspace{-1.2em}
\subsection{Simulation Experiments}
We first conducted comprehensive simulation experiments to demonstrate the effectiveness of AdaptManip, and compared it against the following baselines.

\begin{itemize}
\item \textbf{Pure RL.} The policy is trained with RL with the ground-truth 6D object pose, which corresponds to removing the estimator from our framework.

\item \textbf{Pure RL + FK.} In addition to Pure RL, the policy is provided with the hand positions computed via forward kinematics, providing additional context. 

\item \textbf{Motion Imitation.} The policy follows a predefined grasping motion, similar to recent motion-based methods \cite{peng2018deepmimic}. 

\item \textbf{AdaptManip (Ours).} Our full method combines an RL-based manipulation policy with a recurrent object state estimator.

\item \textbf{Pure Vision.} The policy is trained from scratch with vision and does not rely on an AprilTag for pose detection. 

\item \textbf{Oracle.} The policy is identical to Pure RL but is given perfect ground-truth object state information at test time, providing an upper bound on achievable performance.
\end{itemize}


\replaced[id=SH]{All policies except \textbf{Pure Vision} were trained}{All policies were trained} in IsaacLab without using vision, since training with visual inputs is computationally expensive and typically leads to weaker sim-to-real generalization. During evaluation, the object pose was obtained via an AprilTag in both IsaacLab and MuJoCo except for \textbf{Oracle}. MuJoCo served as an unseen simulator to perform sim-to-sim transfer evaluation and assess generalization beyond the training environment. \added[id=SH]{Each reported entry aggregates 135 trials, covering three box masses, three box dimensions, and three friction coefficients, with five random seeds per configuration.} We decomposed the task into three stages: navigation, grasping, and carrying, as illustrated in Fig.~\ref{fig:overview_exp}, and analyzed the results individually.

\begin{table}[t]
\centering
\renewcommand{\arraystretch}{1.2}
\caption{Performance comparison between AdaptManip and baselines across 135 trials in MuJoCo and IsaacLab.}
\label{table:sim-to-sim}

\resizebox{\columnwidth}{!}{%
\begin{tabular}{lccccc}
\toprule
\textbf{Method} & \textbf{Whole} & \textbf{Stage1} & \textbf{Stage2} & \textbf{Drops} ($\downarrow$) & \textbf{Regrasps} ($\uparrow$) \\

\midrule
\multicolumn{6}{c}{\textbf{IsaacLab}} \\

Pure RL & 
\meanstd{0.62}{0.48} &
\meanstd{0.98}{0.14} &
\meanstd{0.88}{0.32} &
\meanstd{1.79}{2.39} &
\meanstd{5.74}{3.91} \\

Pure RL + FK & 
\meanstd{0.88}{0.32} &
\meanstd{0.93}{0.25} &
\meanstd{0.92}{0.27} &
\meanstd{0.29}{0.89} &
\meanstd{2.94}{1.88} \\

Motion Imitation & 
\meanstd{0.42}{0.46} &
\meanstd{0.96}{0.19} &
\meanstd{0.93}{0.18} &
\meanstd{3.81}{4.77} &
\meanstd{2.94}{1.99} \\

AdaptManip (Ours) &
\meanstd{0.85}{0.35} &
\meanstd{0.97}{0.17} &
\meanstd{0.92}{0.26} &
\meanstd{0.49}{1.14} &
\meanstd{2.12}{1.95} \\

Pure Vision &
\meanstd{0.63}{0.42} &
\meanstd{0.95}{0.22} &
\meanstd{0.91}{0.20} &
\meanstd{12.44}{9.36} &
\meanstd{23.55}{15.32} \\

Oracle & 
\meanstd{0.91}{0.28} &
\meanstd{0.98}{0.14} &
\meanstd{0.97}{0.15} &
\meanstd{0.46}{1.16} &
\meanstd{2.47}{1.48} \\

\midrule

\multicolumn{6}{c}{\textbf{MuJoCo}} \\
Pure RL & 
\meanstd{0.37}{0.48} &
\meanstd{0.84}{0.36} &
\meanstd{0.80}{0.40} &
\meanstd{2.11}{1.01} &
\meanstd{4.20}{3.78} \\

Pure RL + FK & 
\meanstd{0.61}{0.49} &
\meanstd{0.95}{0.22} &
\meanstd{0.93}{0.26} &
\meanstd{2.44}{1.29} &
\meanstd{5.24}{4.53} \\

Motion Imitation & 
\meanstd{0.00}{0.00} &
\meanstd{0.93}{0.26} &
\meanstd{0.69}{0.46} &
\meanstd{0.82}{0.44} &
\meanstd{1.78}{0.93} \\

AdaptManip (Ours) &
\meanstd{0.75}{0.43} &
\meanstd{0.90}{0.30} &
\meanstd{0.88}{0.32} &
\meanstd{1.94}{0.68} &
\meanstd{6.32}{3.95} \\

Pure Vision &
\meanstd{0.37}{0.48} &
\meanstd{0.77}{0.42} &
\meanstd{0.75}{0.43} &
\meanstd{2.09}{0.98} &
\meanstd{4.79}{4.86} \\

Oracle & 
\meanstd{0.79}{0.41} &
\meanstd{0.98}{0.12} &
\meanstd{0.97}{0.16} &
\meanstd{2.16}{1.01} &
\meanstd{6.37}{4.94} \\

\bottomrule
\end{tabular}%
} 
\vspace{-1.75em}
\end{table}

Overall, our results showed that \textbf{AdaptManip} not only achieved competitive performance in IsaacLab (the training environment) but also outperformed the baselines in the unseen MuJoCo environment (Table~\ref{table:sim-to-sim}). In IsaacLab, \textbf{AdaptManip} achieved an 85\% success rate, which was comparable to \textbf{Pure RL + FK} and significantly better than \textbf{Pure RL}, \textbf{Pure Vision}, and the \textbf{Motion Imitation} baseline. Interestingly, \textbf{Pure RL + FK} achieved a success rate of 88\%, substantially outperforming \textbf{Pure RL} at 62\%, which highlighted the importance of informative state representations in reinforcement learning~\cite{kim2021observation}. \textbf{Pure Vision} performed similarly to \textbf{Pure RL}, further showing the relevance of \replaced[id=SH]{the state representation}{state representation and harm from visual occlusion}. In our experiments, \textbf{Motion Imitation} did not perform effectively, as single motion tracking lacked the flexibility required to adapt to diverse interaction scenarios. Overall, the performance gap among \textbf{AdaptManip}, \textbf{Pure RL + FK}, and \textbf{Oracle} was relatively small in this setting, indicating that all three methods achieved near-optimal performance when reliable object state information was available.

\begin{figure}[t]
\begin{center}
\includegraphics[width=\linewidth]{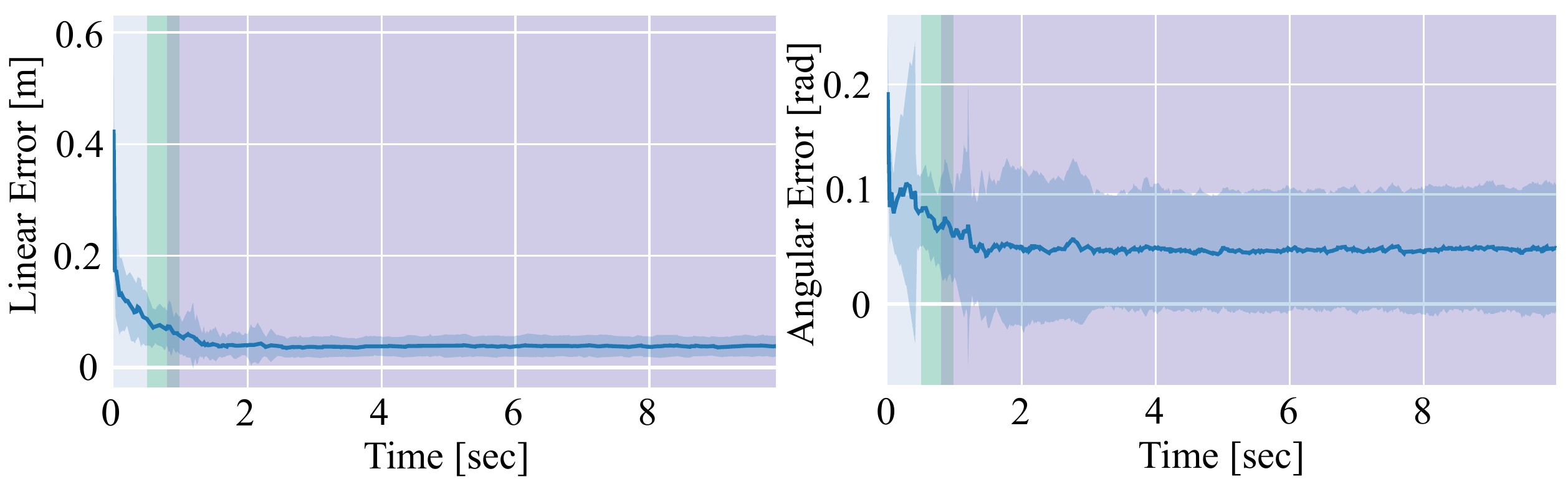}
\end{center}
\caption{State estimation error of our method. Shows mean $\pm$ 1 standard deviation across 50 episodes. The green region shows the area where vision is available, and the purple region shows the area where there is contact between the robot and box.}
\label{fig:pose_error}
\vspace{-1.5em}
\end{figure}

\begin{figure*}[t]
    \centering
    
    \begin{subfigure}{\textwidth}
        \centering
        \includegraphics[width=0.9\textwidth]{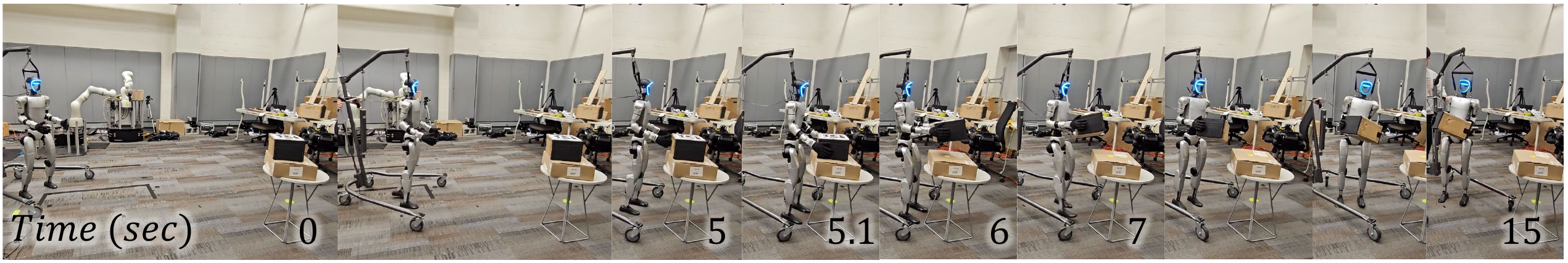}
        \caption{\footnotesize Snapshots of Stage 1 (navigation to the box) and Stage 2 (lifting) during the whole-body loco-manipulation task.}
        \label{fig:locomanipulation_a}
    \end{subfigure}

    \begin{subfigure}{\textwidth}
        \centering
        \includegraphics[width=0.9\textwidth]{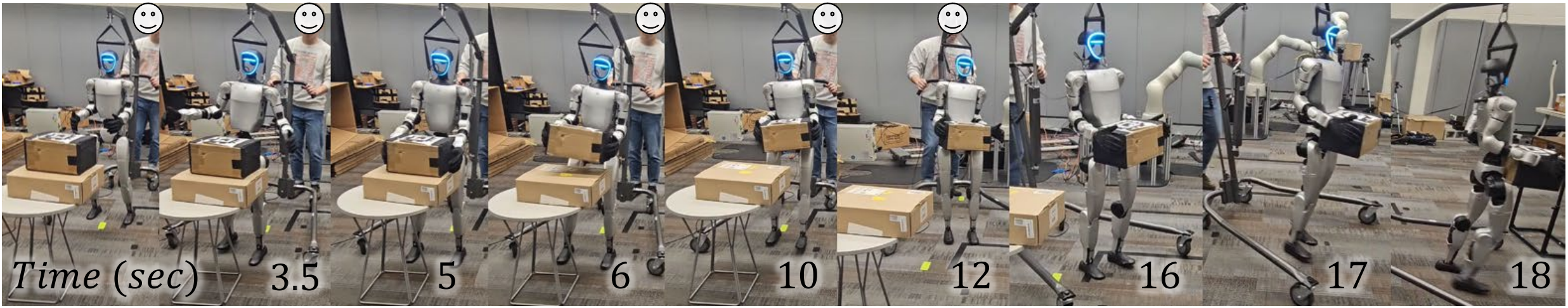}
        \caption{\footnotesize Snapshots illustrating approaching, lifting, and delivery.}
        \label{fig:locomanipulation_b}
    \end{subfigure}

    \caption{Hardware demonstration of the three-stage whole-body loco-manipulation task.}
    \label{fig:locomanipulation}
\vspace{-1.0em}
\end{figure*}

\begin{figure*}[t]
    \centering
    \subfloat[\footnotesize Left and right hand position trajectories during grasping and lifting.]{
        \includegraphics[width=0.48\textwidth]{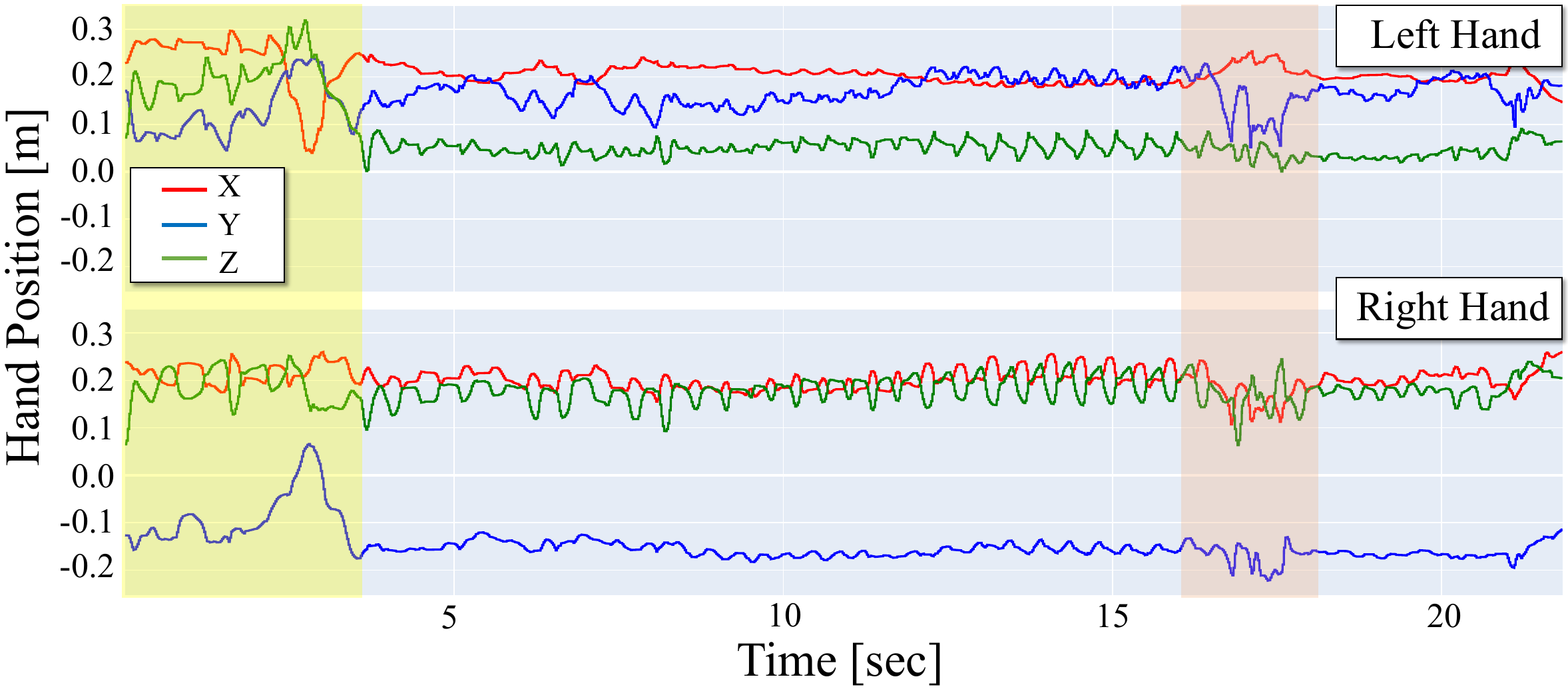}
    }\hfill
    \subfloat[\footnotesize Object position estimates from the recurrent estimator and visual detection.]{%
    \raisebox{0.125ex}{%
        \includegraphics[width=0.48\textwidth]{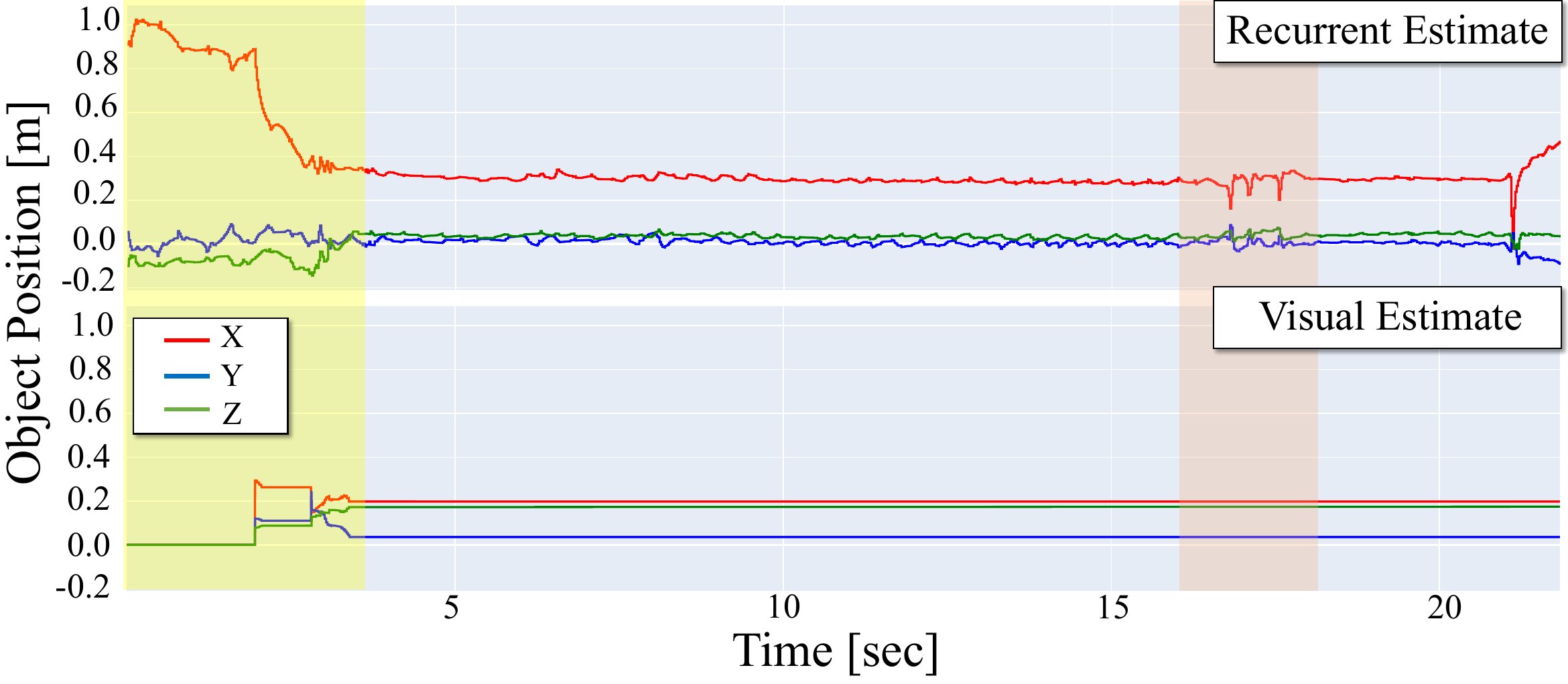}
        }
    }
    \caption{Hardware demonstration of whole-body loco-manipulation. The yellow region indicates the grasp formation phase where the robot carefully coordinates its arms for a secure hold, while the red region highlights the robot's ability to recover from a transient loss of stability through corrective arm motions.}
    \label{fig:locomanip_hw}
    \vspace{-1.5em}
\end{figure*}

However, when we evaluated sim-to-sim transfer, the differences became much more pronounced. \textbf{AdaptManip} achieved a 75\% success rate, which was comparable to the 79\% of \textbf{Oracle}, while also exhibiting a large number of regrasps (6.32), indicating its ability to actively recover from grasping failures. In contrast, \textbf{Pure RL + FK} showed a significantly lower success rate of 61\%, which implied that simply providing end-effector positions was insufficient, as slippage between the hands and the object could not be properly captured. The other three methods, \textbf{Pure RL}, \textbf{Pure Vision}, and \textbf{Motion Imitation}, demonstrated relatively lower success rates, due to the visual sim-to-sim gap and the lack of reliable object state information in the unseen environment. \textbf{Motion Imitation} never completed the task, as a fixed kinematic reference could not handle the change in contact dynamics under sim-to-sim transfer.

We evaluate our policy on unseen objects in IsaacLab under a zero-shot setting. The policy achieves comparable performance across cylinders with different starting orientations (X-axis: $82 \pm 14$, Y-axis: $86 \pm 8$, Z-axis: $78 \pm 11$), while the performance is lower on the sphere ($51 \pm 21$); nevertheless, the result still indicates encouraging generalization to significantly different object geometries.



\vspace{-1.2em}
\subsection{Validation of Object State Estimator}
We conduct further experiments to validate the effectiveness of our learned estimator. For this, we collect 50 episodes of data using our policy under the same randomized conditions used in Table \ref{table:manip-randomization}. For each episode, we collected the ground truth pose information of the box and the input observation information for the estimator, along with flags showing whether the box was visible in the robot camera frame or whether the robot was contacting the box.

We supplied this validation data to our estimator and predicted the pose information. The linear and angular pose errors are shown in Figure \ref{fig:pose_error}. From this, it is clear that initially, there is a relatively large error which is driven down when the AprilTag goes within the FOV of the camera. After that, the error stayed low for the rest of the episode even as the box was carried around due to the information provided from proprioception, which works very well as long as the robot maintains solid contact with the box.


\vspace{-1.15em}
\subsection{Hardware Experiments}
Finally, we deploy the learned policy on a real Unitree G1 humanoid.
As shown in Fig.~\ref{fig:locomanipulation}, our end-to-end, fully autonomous policy completes the task using only onboard sensing and object state estimation. In the grasp formation phase (yellow; Fig.~\ref{fig:locomanip_hw}(a)), the robot carefully coordinates its arms to ensure a secure bimanual grasp before lifting. After grasping, the hand positions remain nearly fixed. This indicates that the robot maintains a stable hold while lifting and transporting the object. Near the later phase of the trial (red), a transient loss of grasp stability occurs and the robot recovers the grasp with a small corrective arm motion.


The object position estimates highlight the advantage of the proposed estimator (see Fig.~\ref{fig:locomanip_hw}(b)). The visual estimate degrades when the object leaves the camera field of view during floating-base motion (e.g., walking). In contrast, our recurrent estimator continues to track the object. This is enabled by jointly leveraging robot proprioception and policy actions, even when visual observations are intermittent.

Overall, these results confirm that our end-to-end policy transfers effectively to real hardware in a zero-shot manner, enabling robust whole-body loco-manipulation using only onboard sensing and object state estimation.


\vspace{-0.7em}
\section{CONCLUSION}
This paper presents a novel framework for completing whole-body, humanoid loco-manipulation tasks. We introduce AdaptManip, a method which combines multi-modal inputs of LiDAR, vision, and proprioception to maintain a recurrent belief of the box pose and hierarchical RL in order to effectively learn a policy which utilizes the pose for picking up and carrying a box from an initial position to a target location. While we show good results for this box lifting task, interesting future work could include trying more extended tasks, incorporating additional sensor modalities for better object state estimation, or using articulated hands for better manipulation.
\vspace{-1.65em}

\section{Acknowledgments}
ChatGPT was used in generating code  and minor editing. All content was reviewed and validated by the authors.

\vspace{-1em}

\bibliographystyle{IEEEtran}
\bibliography{main.bib}

\end{document}